\documentclass[letterpaper, 10 pt, conference]{ieeeconf}  % Comment this line out if you need a4paper

\IEEEoverridecommandlockouts                              % This command is only needed if 
\usepackage{cite}
\usepackage{booktabs}
\usepackage{algorithm}
\usepackage{algpseudocode}
\usepackage{amsmath}
\usepackage{color}
\usepackage{graphicx}
\usepackage{amssymb}
\usepackage[colorlinks,linkcolor=blue,anchorcolor=blue,citecolor=blue]{hyperref}

\title{\LARGE \bf
ReplanVLM: Replanning Robotic Tasks with Visual Language Models
}
\author{Aoran Mei, Guo-Niu Zhu*, ~\IEEEmembership{Member,~IEEE}, Huaxiang Zhang, and Zhongxue Gan*% <-this % stops a space
\thanks{*This work was supported by Shanghai Municipal Science and Technology Major Project under Grant 2021SHZDZX0103 and Key Project of Comprehensive Prosperity Plan of Fudan University under Grant XM06231744. (Corresponding author: Guo-Niu Zhu and Zhongxue Gan)}% <-this % stops a space
\thanks{All authors are with the Academy for Engineering and Technology, Fudan University, Shanghai 200433, China (e-mail: guoniu\_zhu@fudan.edu.cn).}
}

\begin{document}

\maketitle
\thispagestyle{empty}
\pagestyle{empty}

%%%%%%%%%%%%%%%%%%%%%%%%%%%%%%%%%%%%%%%%%%%%%%%%%%%%%%%%%%%%%%%%%%%%%%%%%%%%%%%%
\begin{abstract}
Large language models (LLMs) have gained increasing popularity in robotic task planning due to their exceptional abilities in text analytics and generation, as well as their broad knowledge of the world. However, they fall short in decoding visual cues. LLMs have limited direct perception of the world, which leads to a deficient grasp of the current state of the world. By contrast, the emergence of visual language models (VLMs) fills this gap by integrating visual perception modules, which can enhance the autonomy of robotic task planning. Despite these advancements, VLMs still face challenges, such as the potential for task execution errors, even when provided with accurate instructions. To address such issues, this paper proposes a ReplanVLM framework for robotic task planning. In this study, we focus on error correction interventions. An internal error correction mechanism and an external error correction mechanism are presented to correct errors under corresponding phases. A replan strategy is developed to replan tasks or correct error codes when task execution fails. Experimental results on real robots and in simulation environments have demonstrated the superiority of the proposed framework, with higher success rates and robust error correction capabilities in open-world tasks. Videos of our experiments are available at \textsl{\textcolor[rgb]{0,0,1}{https://youtu.be/NPk2pWKazJc}}.
\end{abstract}

%%%%%%%%%%%%%%%%%%%%%%%%%%%%%%%%%%%%%%%%%%%%%%%%%%%%%%%%%%%%%%%%%%%%%%%%%%%%%%%%
\section{Introduction}
Large language models (LLMs) have received widespread attention for their remarkable reasoning and understanding capabilities and have been applied in various scenarios, such as human-robot interaction \cite{zhang2023large}, action planning \cite{hori2024interactively}, visual target navigation \cite{yu2023l3mvn}, and task planning \cite{guan2024leveraging}. With prompt engineering techniques like chain-of-thought (CoT), LLMs are empowered with powerful reasoning abilities for complex task planning and execution \cite{bao2023smart}. For example, Raman et al. \cite{raman2022planning} proposed a prompt-based strategy for extracting executable plans from an LLM. Their work introduced a corrective re-prompting technique to extract executable corrective actions to achieve the intended goal. Silver et al. \cite{silver2023generalized} used GPT-4 to synthesize Python programs and generate task plans in the planning domain definition language. However, handling the dynamic interaction with the environment remains challenging in robotic task planning. It requires not only a precise perception of environmental information but also the prediction of the causal effects of their actions on the environment. Current LLMs have limitations in understanding the physical state of the world, such as physical forms, spatial locations, and physical properties, which are essential for robot task planning and execution. 

To overcome these challenges, recent studies tend to incorporate external vision models to improve the perception of the environment \cite{zhao2023chat}. Under this framework, LLMs assimilate insights from vision models for task planning. Methods such as HuggingGPT \cite{shen2024hugginggpt} and Tidybot \cite{wu2023tidybot} have adopted this technique. Even though this strategy enables the model to understand the environment better and shows potential in robotic task planning, it suffers from serious issues. The performance of such models depends on external vision models, whereas the information provided by the vision model might be incomplete. 

On the other hand, the development of vision language models (VLMs) has provided a novel solution to the challenges faced by LLMs in environmental interaction \cite{gao2023physically}. VLMs integrate techniques from computer vision and natural language processing to deeply understand text information related to visual content and generate corresponding descriptions. By incorporating visual perception capabilities, VLMs can effectively decode visual cues in task planning, not only capturing environmental information but also predicting the causal impact of each action on the environment. 

Although VLMs outperform LLMs in dealing with complex scenes, long-term task planning remains challenging. It involves predicting multiple future steps, each of which may depend on the outcome of the previous step. The complexity of such scenarios requires the model to have a high degree of adaptability and predictive capability to deal with the uncertainty and diversity of the environment. To address such issues, this study proposes a ReplanVLM, a novel robotic task planning paradigm. This framework is developed based on GPT-4V. An internal error correction mechanism and an external error correction mechanism are presented to enhance the performance of the framework in error detection and correction. A replan strategy is suggested to replan tasks when task execution fails. The main contributions of this paper are summarized as follows.

(1) \textbf{VLM-Enhanced ReplanVLM Framework}: We propose a ReplanVLM, a robotic task planning framework based on VLMs. This framework achieves a deep understanding of the environment and task requirements. The introduction of VLMs not only enables the robot to process visual and textual information but also improves the accuracy and efficiency of task planning and execution.

(2) \textbf{Internal and External Error Correction Mechanisms}: We present an internal error correction mechanism to inspect codes, environments, and task requirements to prevent errors caused by hallucinations or misunderstandings. We put forward an external error correction mechanism that reevaluates the environmental state post-interaction between the robot and its surroundings to guarantee the adaptability of task execution to environmental changes.

(3) \textbf{Experimental Validation and Application}: Multiple experiments on real robots and in simulations are conducted to evaluate the efficacy of the proposed ReplanVLM framework for robotic task planning. Experimental results and comparative analysis show that our framework substantially reduces errors in task execution and enhances the robot's autonomy and adaptability in intricate environments. 

\section{Related work}
LLMs and VLMs are becoming increasingly popular in the field of robotic task planning \cite{zeng2023large, mei2024gamevlm}. Lots of studies were proposed to use LLMs and VLMs to generate task plans \cite{li2023interactive}. For example, Luan et al. \cite{luan2024enhancing} presented a multi-layer LLM to enhance the robot's proficiency in handling complex tasks. A VLM and a LLM were integrated to tackle the challenges associated with task planning and execution. Huang et al. \cite{huang2023voxposer} proposed a VoxPoser framework for extracting affordances and constraints for manipulation tasks in the real world. By leveraging their code-writing capabilities, LLMs were used to interact with VLMs to construct 3D value maps to enable zero-shot task trajectory generation for robots. Zhang et al. \cite{zhang2023grounding} introduced a visually-grounded symbolic planning framework. They used VLMs to detect action failures and verify action affordances towards enabling successful plan execution. Although VLMs can provide visual perception information to LLMs, it might be incomplete. For example, VLMs may only provide object names while ignoring the spatial relationship between objects, which could lead to the failure of robotic task execution or even damage equipment. 

To address this issue, some studies directly used VLMs to generate robotic task plans. Hu et al. \cite{hu2023look} put forward a robotic vision-language planning (ViLA) framework. They utilized LLMs to generate a sequence of actionable steps. By integrating perceptual data directly into the reasoning and planning processes, the ViLA allows robots to understand common sense knowledge in the visual world, including spatial layouts and object attributes. Skreta et al. \cite{skreta2024replan} employed VLMs to understand world state information to empower robots to replan in real-time in the event of action plan failures. Shirai et al. \cite{shirai2023vision} introduced a visual-language interpreter (ViLaIn) framework for robot task planning. They used LLM and VLM to generate problem descriptions and drive symbolic planners in a language-guided framework. Wake et al. \cite{wake2023gpt} presented a multimodal robot task planning pipeline using GPT-4V. By integrating observations from human demonstration, their proposed GPT-4V(ion) succeeded in converting human actions from videos into robot-executable programs. Even if these methods made significant advances in generating task plans, they do not have error correction mechanisms. When VLMs misunderstand instructions or environments, they may generate incorrect plans.

To bridge this gap, this paper proposes a ReplanVLM framework, which introduces internal and external error correction mechanisms to handle internal (e.g., code generation errors) and external errors (e.g., task failures due to environmental changes during execution), respectively.

\section{ReplanVLM: model structure and details}
\subsection{Framework of the proposed ReplanVLM}
\begin{figure*}[ht]
\centering
\includegraphics[width=17cm]{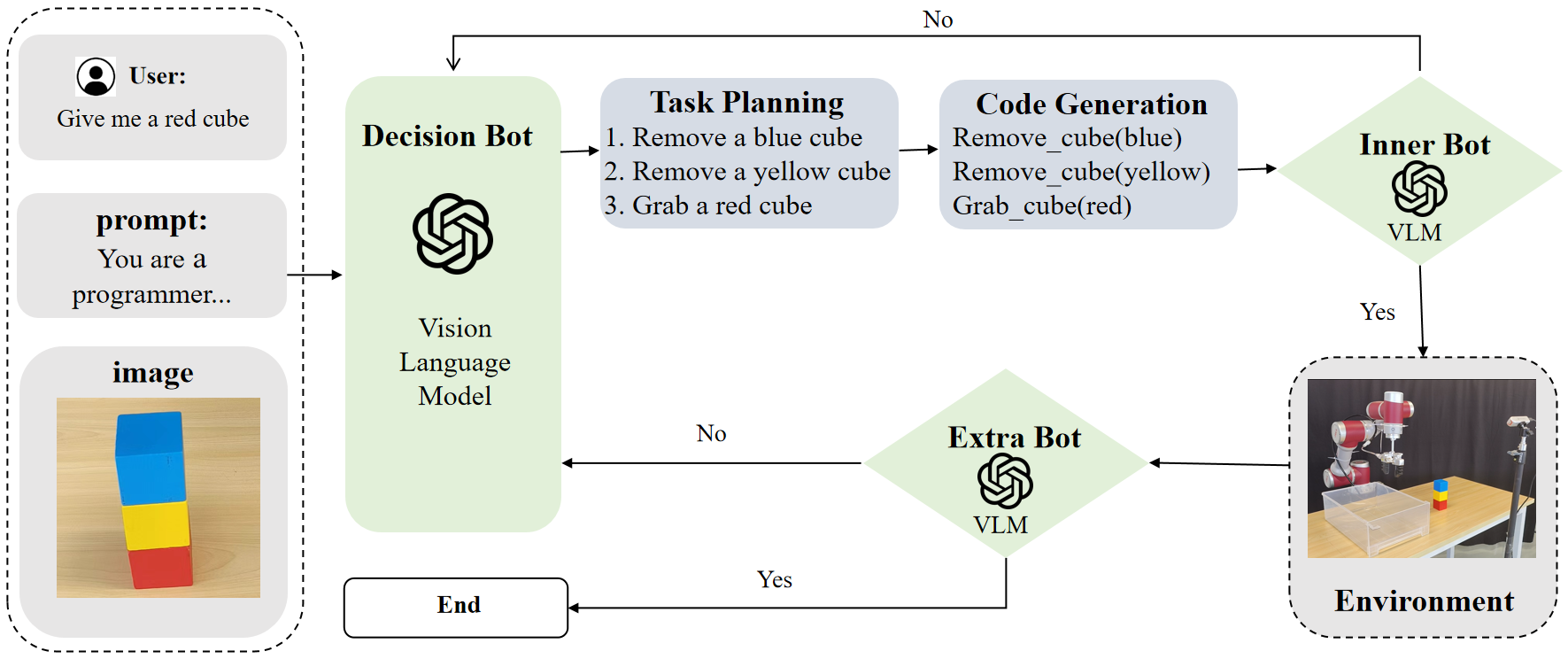}
\caption{ReplanVLM overview. We propose a ReplanVLM framework, which is composed of an internal error correction mechanism and an external error correction mechanism. The internal error correction mechanism refers to an Inner Bot, which is developed to assess and correct the task plans and codes generated by the Decision Bot. The external error correction mechanism involves an Extral Bot, which is used to determine whether the task is completed. If the task is deemed incomplete, the Extra Bot identifies the cause of the failure and sends it back to the Decision Bot to replan the task.}
\label{figure_2}
\end{figure*}

Fig. \ref{figure_2} shows an overview of the ReplanVLM framework, which is developed based on GPT-4V. To start with, the model takes textual instructions as input, e.g., ``help me get an apple." It can also handle abstract requests, like ``I'm a bit thirsty." In this case, the model is capable of making decisions to grab a cup or an apple based on the recognition results of environmental information. In addition to the textual instructions, the input can be a combination of images and text, e.g., ``arrange the blocks in reality according to the sequence shown in the picture." After the initial input stage, a Decision Bot is introduced to plan the task based on the provided input and generate code. The code is then passed to an Inner Bot for review. If the Inner Bot finds errors (i.e., ``No"), such as code mistakes or task planning issues, the Inner Bot will send the reason back to the Decision Bot to regenerate new task plans and codes. Otherwise (i.e., ``Yes"), the robot executes the task plan. Once the execution is completed, an Extra Bot is suggested to compare the current environment with its initial state based on the task requirements. If the task is identified as a failure (i.e., ``No"), the Extra Bot will send the failure information back to the Decision Bot to replan the task. Otherwise (i.e., ``Yes"), the iteration would be terminated. 

This framework is proposed for various tasks and objects even though Fig. \ref{figure_2} uses cubes as examples. In line with the framework, pseudocode (see Algorithm \ref{algorithm-1}) is provided to illustrate the overall process, where $\ell$ represents the tasks to be completed, $\mathbf{t}$ is a certain moment, and $\mathbf{x}_t$ is the state of the environment at time $\mathbf{t}$.

\begin{algorithm}
\caption{\textbf{ReplanVLM}}
\begin{algorithmic}[1]
\State Initial time $\mathbf{t = 0}$, initialize environmental information $\mathbf{x}_t$, initial instructions $\ell$ \Comment{Initialize the required data}
\While{True} 
    \State $\mathbf{p} = \text{VLM}(\mathbf{x}_t, \ell)$ \Comment{Generate task plan}
    \State \text{code} = \text{VLM}($\mathbf{p}$)  \Comment{Generate code }
    \If {$\text{VLM}(\mathbf{x}_t, \ell, \text{code}, \mathbf{p}) == \text{no}$}\Comment{Internal error correction mechanism}
    \State \textbf{t = t + 1}
    \State \textbf{continue}
\EndIf
    \State execute code \Comment{Robots perform tasks}
    \If{$\text{VLM}(\mathbf{x}_t, \mathbf{x}_{\text{new}},\ell) == \text{yes}$} \Comment{External error correction mechanism}
    \State \textbf{break}  \Comment{Task completed}
\Else
    \State \textbf{t = t + 1}
    \State \textbf{continue} \Comment{Task failed}
\EndIf
\EndWhile
\end{algorithmic}
\label{algorithm-1}
\end{algorithm}

\subsection{Internal Error Correction Mechanism}
The internal error correction mechanism is proposed to mitigate the hallucination problem in LLMs during the task generation process. Hallucination is a challenging issue in the process of generating task plans. Incorrect codes may be produced even when the task planning is correct. Another significant role of the internal error correction mechanism is to utilize VLMs to predict the potential impact that their actions might have on the surrounding environment. Consider a scenario where the objective is to ``grab the red block," but a yellow block rests on the red one. In some cases, the model might attempt to grab the red block directly, disregarding the yellow block above it. This action would potentially cause harm to the objects or even damage the robots. To tackle such issues, this study presents an internal error correction mechanism to prevent these problems to enhance the safety and reliability of task execution. The internal error correction mechanism involves an Inner Bot. Based on an internal feedback strategy, the internal error correction mechanism can assess and correct the task plans and codes generated by the Decision Bot. It can be conducted as follows.

\textbf{Format Check:} The Inner Bot first verifies whether the task plans and codes generated by the Decision Bot meet the format requirements. This step prevents the generation of unsuitable text (e.g., blanks) or pseudocode, which can occur even with correct prompts.

\textbf{Matching Check:} Then, Inner Bot checks the alignment of the generated plans and codes. This step helps avoid errors caused by discrepancies between codes and task plans.

\textbf{Plan and Code Verification:} To verify the accuracy of the task plans and codes generated by Decision Bot, Inner Bot regenerates the task plans and codes based on the environment and codebase information, and compares whether the two sets of codes have the same functionality. Such a comparison enhances the framework to detect and tolerate errors.

\begin{figure}[ht]
\centering
\includegraphics[width=8cm]{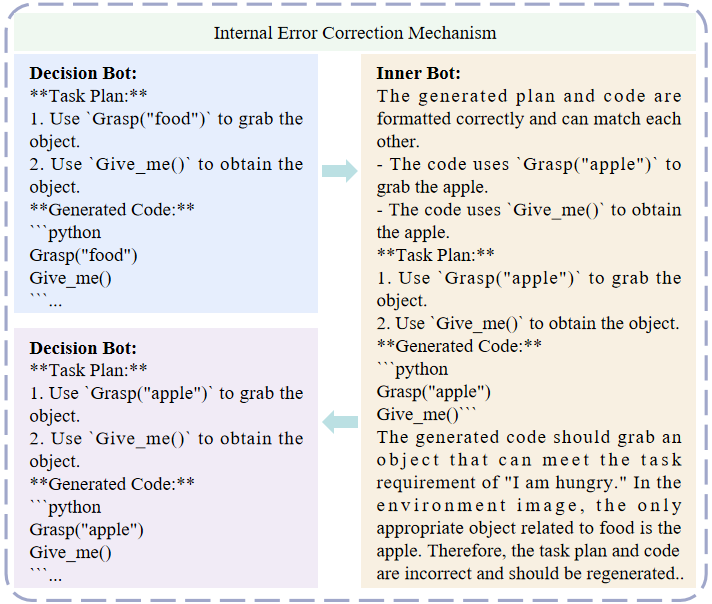}
\caption{An example of the internal error correction mechanism. For the task requirement ``I am hungry" in Task 1, the Decision Bot generates plans and codes to grab ``food." The Inner Bot recommends specifying the object to be grabbed as an ``apple" and then relays this suggestion back to the Decision Bot. Finally, the Decision Bot creates a new and accurate plan.}
\label{Inner_correction}
\end{figure}

If any of the above three points are not satisfied, the feedback generated by the Inner Bot will be passed to the Decision Bot. Given that the Inner Bot's response might contain inaccuracies, this feedback only serves as a reference for the Decision Bot, not a decisive factor. Subsequently, the Decision Bot will regenerate plans and codes based on the known information. If their opinions remain inconsistent after five cycles, the task is deemed a failure. If all three points are satisfied, the robot will execute the task according to the plans and codes generated by the Decision Bot. Fig. \ref{Inner_correction} illustrates an example of the internal error correction mechanism in Task 1 (details can be found in Section \ref{subsec-task}).

\subsection{External Error Correction Mechanism}
The external error correction mechanism is presented to assess the completion status of tasks. During the task execution, robots often encounter unpredictable challenges that can result in task failure. These challenges include but are not limited to, sudden changes in the environment or failures in the robot system. In this context, the external error correction mechanism is crucial. It refers to an Extra Bot, which is utilized to determine whether the task is completed. This mechanism analyzes environmental changes, compares these changes with task requirements, identifies potential causes of task failure, and records such information for task replanning, which is conducted as follows.

The Extra Bot evaluates images of the environment taken before and after task execution. Using these images, along with task requirements, and the task plans and codes generated by the Decision Bot, it assesses whether the task has been successfully completed. If the task is deemed incomplete, the Extra Bot identifies the perceived error reasons and sends them to the Decision Bot. The Decision Bot then uses this feedback and the current environmental information to regenerate task plans and codes. If the Extra Bot confirms task completion, the ReplanVLM framework terminates.

\begin{figure}[ht]
\centering
\includegraphics[width=8cm]{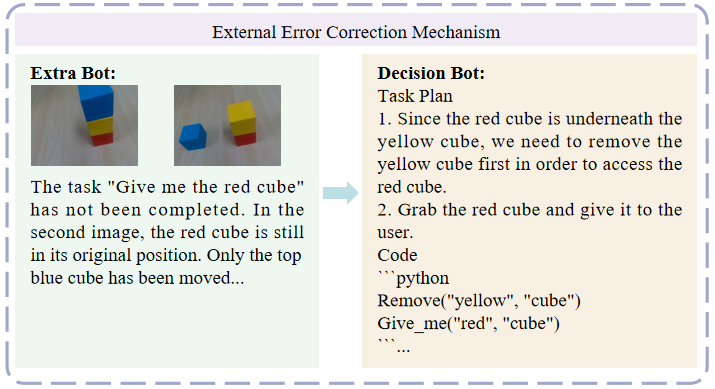}
\caption{An example of the external error correction mechanism. For the task requirement ``give me the red cube" in Task 2, the Extra Bot compares the position of the red cube before and after task execution and finds that it has not moved. Then, the Extra Bot sends this information to the Decision Bot, suggesting that the yellow cube be removed first before grabbing the red cube. Based on this feedback, the Decision Bot generates correct plans and codes.}
\label{External correction}
\end{figure}

By providing such external feedback techniques, the external error correction mechanism can not only enhance the system's ability to understand failure scenarios but also optimize task execution strategies. Fig. \ref{External correction} depicts an example of the external error correction mechanism in Task 2 (details can be found in Section \ref{subsec-task}).

\subsection{Object Detection}
In this study, Yolov8 algorithm is used for object detection. Firstly, it identifies the object's pixel coordinates based on the bounding box in RGB images. Then, the object's pose information is gained by integrating these pixel coordinates and the depth information. Besides, this algorithm has tracking capabilities. It can accurately detect objects and give their pose information even if they move. In some tasks, such as ``give\_me()," the movement endpoints are predetermined. While in other tasks, like ``put(`plate')," the robot dynamically chooses a suitable placement based on environmental settings.

\section{Experiments}
\subsection{Experimental Platform}
\begin{figure}[ht]
\centering
\includegraphics[width=8cm]{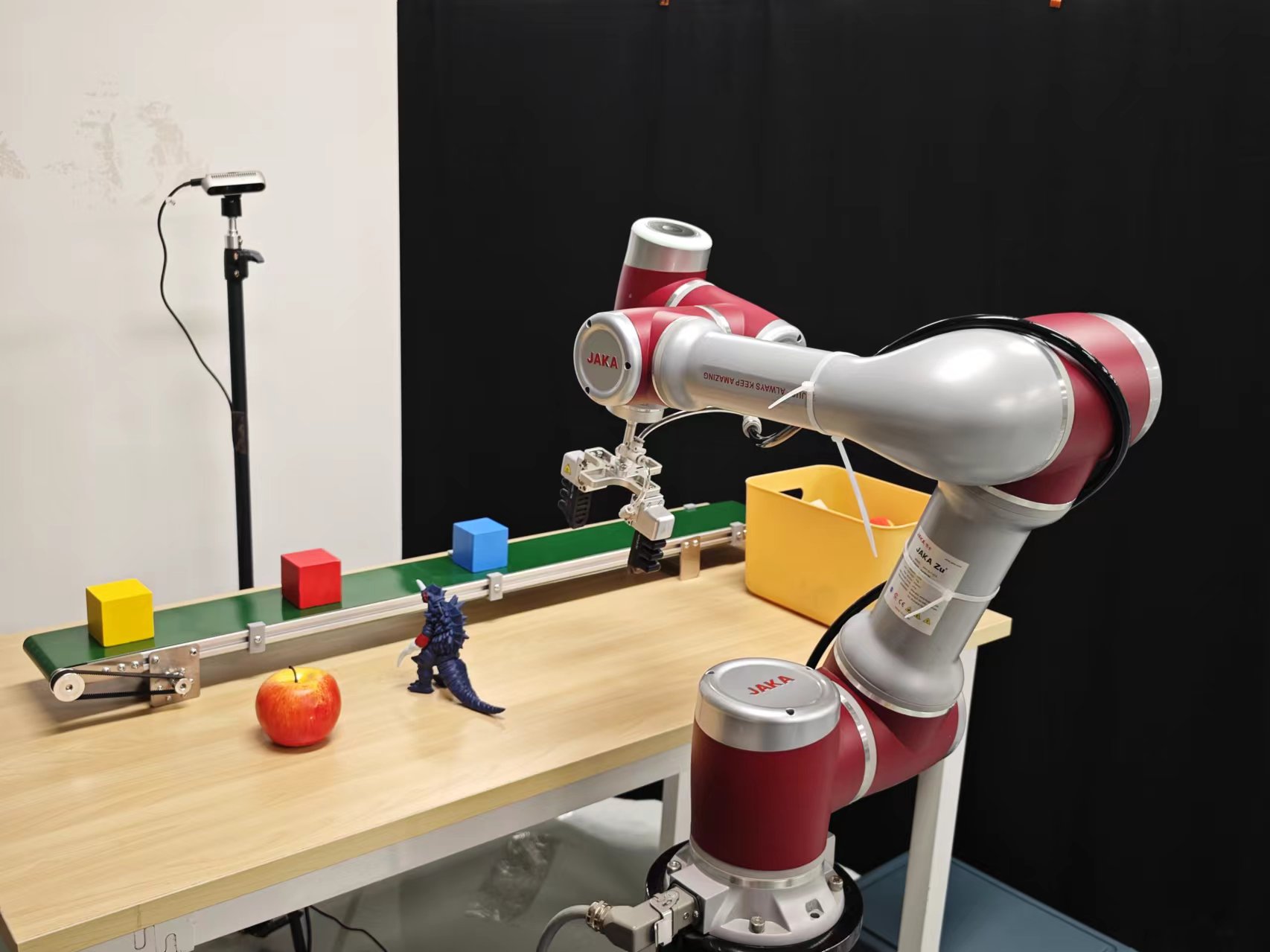}
\caption{Overview of the system setup.}
\label{figure_3}
\end{figure}

As illustrated in Fig. \ref{figure_3}, the experimental setup consists of a 6 degrees of freedom robotic arm (i.e., JAKA Zu 7), an Intel Realsense D435i depth camera, a Rochu pneumatic gripper, and a conveyor belt. A couple of toys and simulated fruits are introduced as objects for the experiment.

\subsection{Prompt}
\begin{figure*}[ht]
\centering
\includegraphics[width=16cm]{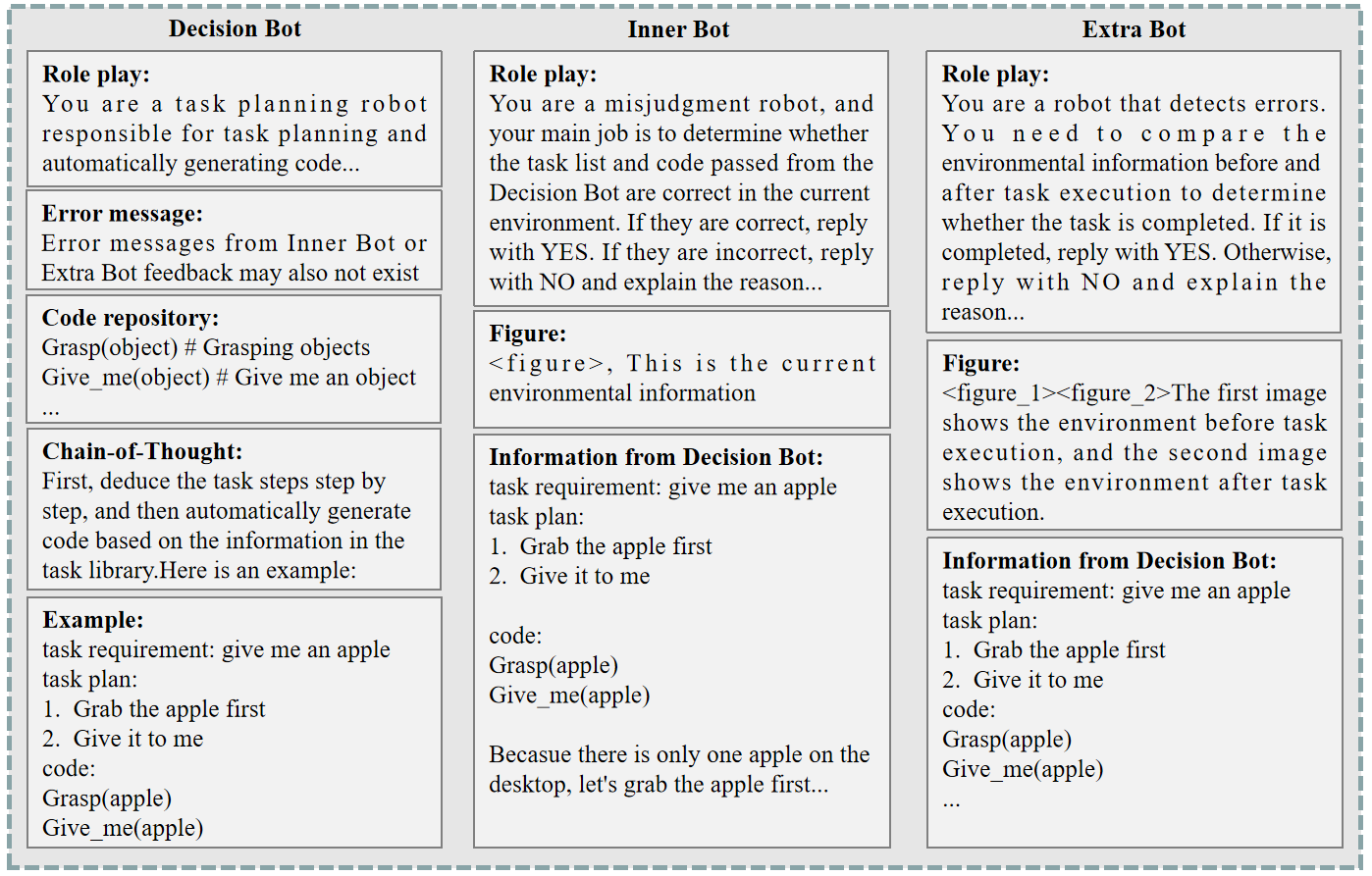}
\caption{Example prompts in the ReplanVLM framework.}
\label{figure_4}
\end{figure*}

For each component of the ReplanVLM framework, we propose different prompts. As shown in Fig. \ref{figure_4}, the prompt of the Decision Bot consists of role-playing, error messages, code repository, CoT, and examples. By contrast, the prompts of the Inner Bot and Extra Bot comprise role-playing, images, and information from the Decision Bot. For each task undertaken by these components, the prompts would vary slightly to accommodate the specifics of the task, e.g., the code repository and role-playing.

\subsection{Task Design}\label{subsec-task}
A series of tasks are proposed to evaluate the effectiveness of the ReplanVLM framework on real robots and in simulation environments. As shown in Table \ref{table-1}, these tasks are designed based on the following criteria. 

\textbf{Semantic Understanding (SU):} the ability to comprehend and interpret the meaning of a language in a way that is akin to human understanding. For example, given the task ``I'm a little hungry," the robot can identify from environmental clues that it should grab an apple. 

\textbf{Spatial Relationship (SR):} the capacity to understand the relative spatial position between objects, such as the ability to stack blocks in order according to a given picture. 

\textbf{External Spatial Constraints (EC):} the ability to recognize various constraints present in the environment. For instance, in the task ``Help me grab the red block," the robot needs to remove the blue and yellow blocks located above the red one. 

\textbf{Understanding of Special Attributes (UA):} the ability to comprehend the attributes of objects, such as ``kind," ``evil," ``dangerous," ``safe," ``edible," etc. 

\textbf{Adaptability to Dynamic Scenes (AS):} assessing the framework's adaptability to dynamic scenes.

\textbf{Minimum Operation Steps (MS):} the minimum operation steps (no error correction process) include robotic arm movement, gripper closure, and gripper opening.

\begin{table}[ht]
\caption{Characteristics of the experimental task}
\begin{center}
\begin{tabular}{*{7}{c}}
\toprule
Task & SU & SR & EC & UA & AS & MS\\
\midrule
Task 1 & \checkmark &  &   & \checkmark & & 6 \\
Task 2 &  & \checkmark & \checkmark  &  & & 21 \\
Task 3 & \checkmark & \checkmark & \checkmark  &  & &14 \\
Task 4 &  & \checkmark &   &  & \checkmark & 22 \\
Task 5 &  & \checkmark &   &  & \checkmark & 13\\
Task 6 &  & \checkmark & \checkmark  &  &  & 13\\
Task 7 &  &  &   & \checkmark & & 7 \\
\bottomrule
\end{tabular}
\label{table-1}
\end{center}
\end{table}

Based on these criteria, seven tasks are proposed as follows.

\textbf{Task 1: Food Demand Recognition.} In this task, a variety of objects, including food, are placed on a table. This task aims to assess whether the robot can accurately understand the user's needs upon receiving the instruction ``I'm hungry" and successfully identify and grasp the apple on the table.

\textbf{Task 2: Grasping the Block.} In this task, three colored blocks are arranged in a specific order—blue on top, yellow in the middle, and red at the bottom. Before grabbing the red block and placing it into a cardboard box, the robot needs to remove the blue and yellow blocks.

\textbf{Task 3: Stacking Blocks.} By providing the VLM with a picture showing the order of block stacking, the robot is required to stack the blocks in the same order.

\textbf{Task 4: Sequential Arrangement on a Conveyor Belt.} This task requires the robot to reorder the objects in a designated order on a moving conveyor belt. This task aims to examine the robot's response to dynamic environments, especially its ability to monitor and adjust the arrangement of objects in dynamic scenarios.

\textbf{Task 5: Categorization and Transport.} In this task, the robot is required to classify the fruits on the conveyor belt and place them into different plates. This task intends to test the robot's ability to recognize and transport items, assessing its efficiency in handling objects with various physical properties.

\textbf{Task 6: Grabbing Invisible Objects.} This task requires the robot to identify and grab a block inside a drawer. Since the visual detection system cannot directly identify the block, the robot must open the drawer and then grab the object.

\textbf{Task 7: Toy Recognition.} In this task, the robot is asked to identify and grab toys with given attributes, e.g., grabbing a toy with evil attributes. This task aims to evaluate the VLM's ability to recognize intricate features, such as moral attributes.

In the experiment, the robot runs ten rounds on each task. The average success rate of the ten rounds will be taken as the evaluation metric to indicate the system performance on the corresponding task.

\subsection{Ablation Experiment}\label{subsec-ablation}
To evaluate the relative importance of each module within the ReplanVLM framework, ablation studies are conducted to assess the contribution of the internal and external error correction mechanisms on the overall performance of the ReplanVLM. Accordingly, four models are developed in the ablation experiment, i.e., the ReplanVLM, the ReplanVLM without internal error correction mechanism (ReplanVLM-internal), the ReplanVLM without external error correction mechanism (ReplanVLM-external), and the ReplanVLM without error correction mechanisms (ReplanVLM-both). Each model will be tested on the tasks as presented in \ref{subsec-task}.

\subsection{Error Correction Experiment}
Furthermore, error correction experiments are introduced to test the robustness of the ReplanVLM framework when faced with external disturbances. Specifically, this experiment focuses on the model's ability to detect and correct errors after a failed operation. In this experiment, intentional manual interference is applied to cause a failure during the robot's grabbing task. Then, the robot needs to determine whether the task is completed. If the task is deemed incomplete, a correction mechanism is triggered. An error detection rate and an error correction rate are introduced as evaluation metrics in this experiment. Likewise, the experiment will be run in ten rounds to get statistical metrics.

\subsection{Simulation Experiment}
\begin{figure*}[ht]
\centering
\includegraphics[width=17cm]{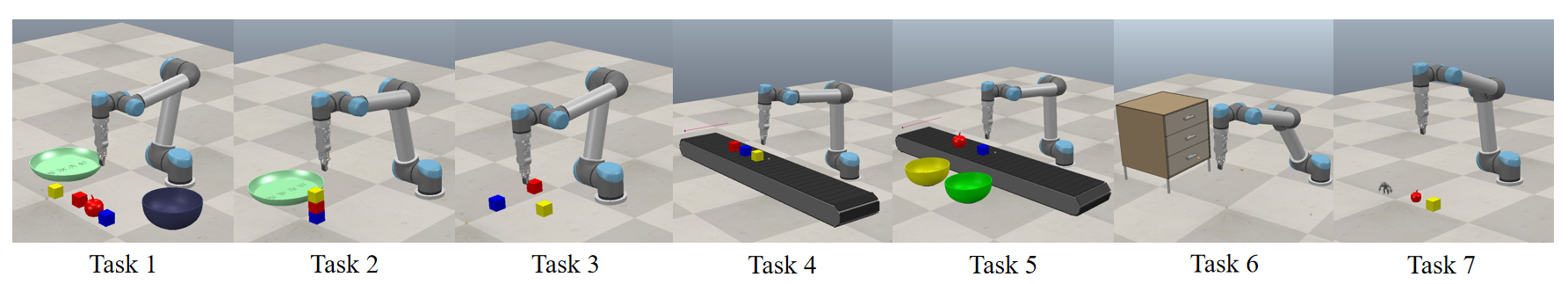}
\caption{Experimental setup in the simulation environment.}
\label{simu_environment}
\end{figure*}

Besides the experiments on real robots, simulations are conducted to evaluate the ReplanVLM framework. CoppeliaSim robotic simulator is introduced to perform the seven tasks described in \ref{subsec-task}. As depicted in Fig. \ref{simu_environment}, a UR5 robotic arm and a RG2 gripper are set up to execute these tasks in the simulation environment.

\section{Results and discussion}
\subsection{Experimental Results}
Table \ref{table-results} illustrates the experimental results on real robots and in simulations. As shown in Table \ref{table-results}, the proposed ReplanVLM achieves an average success rate of 94.2\% on real robots. The success rate can even reach up to 100\% in the task of food demand recognition (Task 1), sequential arrangement on a conveyor belt (Task 4), categorization and transport (Task 5), and toy recognition (Task 7). In task 3 (visual assistance in block stacking), the ReplanVLM framework performs the worst, with a success rate of 80\%. In most of the tasks, the success rate is higher than 90\%. Overall, the ReplanVLM performs well across various scenarios, especially in tasks that require the recognition of complex attributes, the discovery of potential constraints, and the use of world knowledge. The performance across diverse tasks demonstrates the effectiveness of the proposed ReplanVLM in robot task planning.

\begin{table}[ht]
\caption{Experimental results on real robots and in simulations}
\begin{center}
\begin{tabular}{*{3}{c}}
\toprule
Task & Real robots & Simulations \\
\midrule
Task 1 & 100\% & 100\% \\
Task 2 & 90\% & 100\% \\
Task 3 & 100\% & 90\% \\
Task 4 & 80\% & 90\% \\
Task 5 & 100\% & 100\% \\
Task 6 & 90\% & 90\% \\
Task 7 & 100\% & 90\% \\
Average & 94.2\% & 94.2\% \\
\bottomrule
\end{tabular}
\label{table-results}
\end{center}
\end{table}

In addition, the proposed framework achieves an average success rate of 94.2\% in simulations, which is identical to the results on real robots. Specifically, the success rate of Task 1, Task 2, and Task 5 can reach up to 100\%, while the other four tasks get a success rate of 90\%. Therefore, the ReplanVLM framework not only performs well in a single task but also maintains a high level of reliability in multiple scenarios in simulated environments. Moreover, the results of the ReplanVLM framework show a high consistency in both real-world and simulated environments.

\subsection{Comparative Experiment Analysis}
To evaluate the performance of the proposed ReplanVLM, SayCan \cite{brohan2023can} and ProgPrompt \cite{singh2023progprompt} are introduced as baseline methods. As listed in Table \ref{table1}, the proposed framework achieves an average success rate of 94.2\%, while SayCan gets 10\% and ProgPrompt gains 31.4\%. Obviously, the average success rate of our framework is much higher than baseline methods. The proposed framework outperforms baseline methods on all testing tasks. Notably, both the SayCan and ProgPrompt obtain nearly zero success rates in tasks that require a deep understanding of visual information, such as scene analysis with visual constraints and handling objects with complex attributes (e.g., distinguishing between monsters and Ultraman toys). Compared with baseline methods, the proposed framework shows great potential in intricate environments, especially in scenarios that require scene understanding and decision-making capabilities.

\begin{table}[ht]
\caption{Results of comparative studies}
\begin{center}
\begin{tabular}{*{4}{c}}
\toprule
Task & SayCan \cite{brohan2023can} & ProgPrompt \cite{singh2023progprompt} & Ours \\
\midrule
Task 1 & 50\% & 70\% & 100\%   \\
Task 2 & 0\% & 0\% & 90\%   \\
Task 3 & 0\% & 0\% & 100\%   \\
Task 4 & 20\% & 50\% & 80\%   \\
Task 5 & 0\% & 70\% & 100\%   \\
Task 6 & 0\% & 20\% & 90\%   \\
Task 7 & 0\% & 10\% & 100\%   \\
Average & 10.0\% & 31.4\% & 94.2\%   \\
\bottomrule
\end{tabular}
\label{table1}
\end{center}
\end{table}

Given the poor performance of baseline methods, the shortcomings lie in two aspects. Firstly, previous works mainly use a single-modal LLM for robot task planning. Their understanding of the environment mainly relies on external models, which makes it difficult to consider the constraints in the environment. Secondly, previous works do not have effective error detection and correction mechanisms. They cannot regenerate task plans based on previous failed experiences. By contrast, this study introduces a VLM for task planning and proposes internal and external error correction mechanisms to correct errors. Compared with previous works, our work can better understand the environment and regenerate task plans based on failed experiences. The success rate of the task planning is substantially enhanced.

%\begin{figure}[ht]
%\centering
%\includegraphics[width=8 cm]{fig-results.pdf}
%\caption{Experimental setup in the simulation environment.}
%\label{fig_results}
%\end{figure}

\subsection{Ablation Experiment Analysis}
Corresponding to the protocol of the ablation study discussed in \ref{subsec-ablation}, the results of the ablation experiment are given in Table \ref{table2}. As illustrated in Table \ref{table2}, the average success rates of the four methods are obtained as 94.2\% (ReplanVLM), 87.1\% (ReplanVLM-internal), 78.5\% (ReplanVLM-external), and 68.6\% (ReplanVLM-both). Specifically, the ReplanVLM outperforms the other three methods on all testing tasks. With the internal and external error correction mechanisms, the ReplanVLM achieves the highest average success rate. On the other hand, the average success rate of the ReplanVLM-internal is slightly lower than the ReplanVLM. Although the ReplanVLM-internal performs as well as the ReplanVLM on some tasks, it lags behind ReplanVLM on most tasks. In addition, the average success rate of the ReplanVLM-external is even lower than the ReplanVLM-internal, while the ReplanVLM-both performs the worst.

\begin{table*}[ht]
\caption{Results of the ablation experiment}
\begin{center}
\begin{tabular}{l *{8}{c}}
\toprule
Method & Task 1 & Task 2 & Task 3 & Task 4 & Task 5 & Task 6 & Task 7 & Average \\
\midrule
ReplanVLM & 100\% & 90\% & 100\% & 80\% & 100\% & 90\% & 100\% & 94.2\% \\
ReplanVLM-internal & 90\% & 90\% & 100\% & 60\% & 90\% & 90\% & 90\% & 87.1\% \\
ReplanVLM-external & 90\% & 80\% & 100\% & 80\% & 80\% & 80\% & 70\% & 78.5\% \\
ReplanVLM-both & 90\% & 60\% & 90\% & 60\% & 80\% & 60\% & 40\% & 68.6\% \\
\bottomrule
\end{tabular}
\label{table2}
\end{center}
\end{table*}

From the results of ablation experiments, it can be concluded that both the internal and external error correction mechanisms contribute substantially to the overall performance of the ReplanVLM framework. By removing any one of the error correction mechanisms, it will lead to a significant decrease in the success rate. By comparison, the impact of the external error correction mechanism on the overall framework is more pronounced than the internal error correction strategy. 

\subsection{Error Correction Experiment Analysis}
\begin{figure}[ht]
\centering
\includegraphics[width=8.5cm]{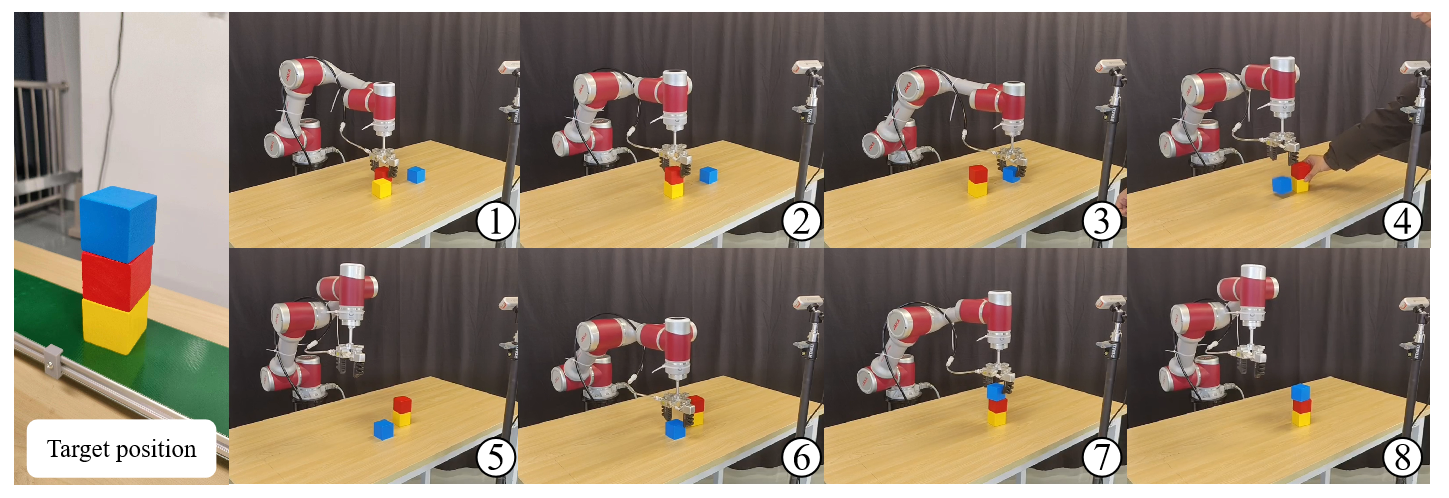}
\caption{Correction process. Taking Task 3 as an example, the image on the left shows the final position of the blocks. During the process of the robot placing the blue block, manual interference resulted in the blue block not being placed on top of the red block (Step 4). Subsequently, the VLM detected and corrected the error, ultimately completing the task.}
\label{figure_6}
\end{figure}

Table \ref{table3} shows the results of the error correction experiment. As listed in Table \ref{table3}, the proposed framework gains an average error detection rate of 95.7\%, with an average error correction rate of 80\%. The error detection rate can reach up to 100\% in the majority of tasks. Fig. \ref{figure_6} shows an example of the correction experiment. As it can be seen from the experimental results, the ReplanVLM framework can effectively detect errors and correct them during the task execution. Although there is a fluctuation in error correction rate across different tasks, the proposed framework shows strong potential in enhancing the accuracy and stability of robot task execution, especially in intricate environments.

\begin{table}[ht]
\caption{Results of the error correction experiment}
\begin{center}
\begin{tabular}{*{3}{c}}
\toprule
Task & Error detection rate & Error correction rate \\
\midrule
Task 1 & 100\% & 90\% \\
Task 2 & 100\% & 80\% \\
Task 3 & 90\% & 60\% \\
Task 4 & 80\% & 60\% \\
Task 5 & 100\% & 100\% \\
Task 6 & 100\% & 80\% \\
Task 7 & 100\% & 90\% \\
Average & 95.7\% & 80.0\% \\
\bottomrule
\end{tabular}
\label{table3}
\end{center}
\end{table}

\subsection{Discussion}
With the aid of the internal and external error correction mechanisms, the proposed ReplanVLM framework demonstrates an impressive performance on real robots and in simulation environments. It outperforms baseline methods in the comparative studies, with a better performance on all testing tasks. Either the internal error correction mechanism or the external error correction mechanism contributes significantly to the overall performance of the ReplanVLM framework.

Despite the prominent performance, this study has some limitations. First of all, this study only uses text and visual information for robot task planning. It cannot perceive other modal information such as touch and hearing. Secondly, although the LLM possesses broad world knowledge, it is difficult to handle complex manipulation tasks, such as cooking, through simple programming. Thirdly, the success rate of the task planning and execution still suffers from certain errors, such as the limitations in the VLM's visual processing and defects with the error correction mechanisms. In addition, the network latency caused by using the GPT-4V deteriorates the real-time performance of the overall system. The framework cannot apply to real-time systems.

\section{Conclusions}
This paper proposed a ReplanVLM framework to enhance the success rate of robots in task planning and execution. By incorporating VLM techniques in robotic task planning, the ReplanVLM framework introduces internal and external error correction mechanisms that can effectively identify and rectify errors encountered during task execution. Extensive experiments in real-world and simulated environments were conducted to evaluate the effectiveness and robustness of the proposed method in various scenarios. Experimental results and comparative analysis demonstrate that the ReplanVLM framework can reliably detect and correct errors in diverse situations. The success rate of task planning and execution is substantially enhanced.

In the future, we will introduce some other perceptual information, such as tactile information, to enhance the framework to better understand the environment so that the task plans can be generated more accurately. In addition, we will investigate the potential of multimodal LLMs in completing complex manipulation tasks. 

%%%%%%%%%%%%%%%%%%%%%%%%%%%%%%%%%%%%%%%%%%%%%%%%%%%%%%%%%%%%%%%%%%%%%%%%%%%%%%%%
\bibliographystyle{IEEEtran}
\bibliography{ref}
\end{document}